\documentclass[12pt]{article}
\usepackage{hyperref}
\usepackage{graphicx}
\usepackage{float}
\usepackage{setspace}
\usepackage{lineno}
\usepackage{amsmath}
\usepackage{amssymb}
\usepackage{subcaption}
\usepackage[numbers, square, sort&compress]{natbib}
\usepackage{authblk}
\usepackage{geometry}
\usepackage{tabularx}
\usepackage{booktabs}
\usepackage{multirow}
\usepackage{array}
\usepackage{makecell}
\usepackage{longtable}
\usepackage{pdflscape}
\usepackage{afterpage}
\usepackage{rotating}
\usepackage{lscape}
\usepackage{array}
\usepackage{tabularx}
\usepackage{graphicx}
\usepackage{color, soul}
\usepackage{listings}
\usepackage{xparse}

\begin{document}


\title{Using Generative Text Models to Create Qualitative Codebooks for Student Evaluations of Teaching}

\author{Andrew Katz\\
\small{Department of Engineering Education}\\
\small{Virginia Tech}
\and
Mitch Gerhardt\\
\small{Department of Engineering Education}\\
\small{Virginia Tech}
\and
Michelle Soledad\\
\small{Department of Engineering Education}\\
\small{Virginia Tech}}

\date{}

\maketitle

\begin{abstract}
Feedback is a critical aspect of improvement. Unfortunately, when there is a lot of feedback from multiple sources, it can be difficult to distill the information into actionable insights. Consider student evaluations of teaching (SETs), which are important sources of feedback for educators. They can give instructors insights into what worked during a semester. A collection of SETs can also be useful to administrators as signals for courses or entire programs. However, on a large scale as in high-enrollment courses or administrative records over several years, the volume of SETs can render them difficult to analyze. In this paper, we discuss a novel method for analyzing SETs using natural language processing (NLP) and large language models (LLMs). We demonstrate the method by applying it to a corpus of 5,000 SETs from a large public university. We show that the method can be used to extract, embed, cluster, and summarize the SETs to identify the themes they express. More generally, this work illustrates how to use the combination of NLP techniques and LLMs to generate a codebook for SETs. We conclude by discussing the implications of this method for analyzing SETs and other types of student writing in teaching and research settings.

\end{abstract}


\section{Introduction}

Text and speech constitute essential elements of qualitative data utilized across various research domains \cite{miles_qualitative_2014,pope2000qualitative}. These data, including interviews, transcripts, reports, social media posts, written answers to open-ended response questions, and essays, are traditionally analyzed through methods such as qualitative thematic analysis \cite{terry_sage_2017}. However, these analytic approaches often necessitate significant research investment, incurring costs for both projects and researchers. For instance, conventional tools like MAXQDA \cite{radiker2020focused,gizzi2021practice} and NVivo \cite{edwards2014qualitative} come with price tags and demand extensive research hours and revisions \cite{basit2003manual}. Moreover, as the data grow so does the time needed for analysis. Despite these challenges, thematic analysis remains prevalent in qualitative research due to its capacity to extract detailed insights and themes \cite{miles_qualitative_2014,pope2000qualitative,terry_sage_2017}. Nonetheless, the relationship between data size and researcher time is nonlinear, making large and existing datasets increasingly difficult to manage. Additionally, datasets containing ill-formatted or unstructured data pose challenges to traditional qualitative analysis due to their incompatibility with standard tools or varying datatype content \cite{patibandla2015unstructured}.

During qualitative thematic analysis, researchers scrutinize data to extract or identify salient themes and concepts, referred to as ``codes.'' This process involves applying codes based on choices made by researchers, such as whether to include a theoretical stance or employ an inductive or deductive coding approach \cite{terry_sage_2017}. Inductive coding, a ``bottom-up'' approach, derives codes (and resulting themes) from the data itself, allowing the data to dictate the identification of meaning and interpretation. Conversely, deductive coding involves applying existing theoretical concepts or theories as a lens to interpret the data - a ``top-down'' approach. Thus, inductive and deductive processes differ in the extent to which the data themselves are dictating the contents of the ``codebook''. Nevertheless, both processes are subject to researcher subjectivity and positional bias, leading to the common suggestion that multiple researchers should analyze the same data \cite{miles_qualitative_2014, terry_sage_2017}.

Qualitative data hold substantial research potential given the density of the information they contain but entail costs and time-consuming processes. Fortunately, recent advancements in natural language processing (NLP), a subset of machine learning, integrate large language models (LLMs) to automate language understanding and synthesis \cite{crowston_machine_2010,katz_utility_2023}. These tools offer significant potential for enhancing the accuracy and efficiency of qualitative thematic analysis \cite{garman_developing_2021, bhaduri_nlp_2017}. Current NLP applications leverage LLMs to analyze sentiment, synthesize text, design feedback systems, develop tutoring systems for computer education, and model stance in student essays \cite{kastrati_sentiment_2021,somers_applying_2021,shaik_review_2022,sunar_natural_2024,ganesh_response_2022,mathew_nlp_based_2021,chiu_visualizing_2023,persing_modeling_2016}. Despite these applications, there remains considerable potential for NLP-directed qualitative data analysis to more closely align with current human-based approaches.

In this paper, we aim to demonstrate the development and utilization of an NLP-based method for inductive qualitative data analysis. We demonstrate this method on a set of 5,000 student evaluations of teaching (SETs) containing student feedback for instructors. Our method generates a codebook that summarizes the prevalent themes and concepts within these responses. We show that this generated codebook mirrors the codebook produced from a previously completed traditional thematic analysis on a selection of the same dataset \cite{soledad_leveraging_2017, soledad_understanding_2019}. While previous approaches have utilized NLP for deductive qualitative coding \cite{tai_examination_2024, katz_exploring_2023}, we present a novel method for inductive thematic coding using a robust, iterative process that extracts, embeds, clusters, and summarizes qualitative data into a codebook akin to traditional inductive coding processes. This automated, data-secure method employs open-source LLMs, avoiding proprietary models, like OpenAI's GPT models and Google's Gemini models that may utilize user data for training \cite{noauthor_privacy_2023,noauthor_gemini_nodate}. Additionally, our programmatic approach utilizes open-source APIs deployed locally on one's computer to automate the coding process without the need for user interfaces like chatbots. Consequently, we anticipate researchers can employ our approach to conduct meaningful qualitative analysis securely, efficiently, and expediently. When combined with researcher oversight, there is significant potential for this approach to broaden the scope of qualitative data analysis. 

\section{Literature Review}
\label{sec:lit_review}

 
This section provides insight into the use of NLP techniques for analyzing textual data. We explore the growing application of NLP in education research, including its use in analyzing educational content, generating course materials, and assisting learning processes. Further, we describe SETs and their use in engineering education as an example use case, which were used to validate the methodology. We aim to establish the theoretical and methodological foundations for using NLP in analyzing SETs in education by synthesizing these research aspects. Furthermore, we suggest that this methodology could be applied to similar qualitative data beyond this context, which we will explore in a subsequent section.

\subsection{NLP Use in Education}

Machine learning (ML) continues to be a growing interest within research and industry. As ML technology has advanced, so too have the applications leveraging its unique pattern recognition and decision-making capabilities \cite{alpaydin2021machine,el_naqa_what_2015}. Like many sectors, education is ripe for transformation from ML-driven tools. From their capacity to analyze textual data to insights from unstructured datasets like videos and audio, ML has immense potential to broadly disrupt learning experiences, instruction methods, and educational systems \cite{katz_utility_2023, shaik_review_2022,sunar_natural_2024}.

Natural language processing (NLP), a sub-field of ML, entails designing algorithms and models that enable computers to understand, interpret, and generate human language data \cite{chowdhary_natural_2020,katz_utility_2023,bhaduri_nlp_2017,gillioz_overview_2020}. Modern NLP tasks expand beyond single words, allowing computers to approximate an understanding of the context of words. Consequently, NLP models can classify sentences, generate text content, and summarize text, among other tasks. Here, ``models" refer to the sets of rules or parameters that map input data to output predictions \cite{chowdhary_natural_2020, gillioz_overview_2020}. Simply, a model takes in information (the input) and produces a result or prediction (the output). The model learns from examples in its training data, adjusting its internal parameters to improve its predictions over time. Many NLP models are often highly complex, requiring immense training data and tuning to achieve desired results. However, once trained, these models can be leveraged for a variety of use cases and analyze or generate text beyond what they were trained on \cite{chowdhary_natural_2020, gillioz_overview_2020}. Of course, these models are limited and may contain biases or exhibit unexpected behavior, requiring deliberate and careful use when employed \cite{vig_investigating_2020, arthurs_whose_2020, cotton_chatting_2023, qadir_engineering_2022}.

More recently, NLP models have become more effective at generating text given an input prompt \cite{min_recent_2023}. As many people have experienced with chatbots like ChatGPT \cite{ansari_mapping_2023, cooper_examining_2023}, NLP is used to understand user queries, extract relevant information, and generate appropriate responses. This might involve tasks like entity recognition to identify key information, sentiment analysis to gauge the user's mood feelings toward specific aspects, and language generation to craft natural-sounding replies – all carefully designed processes that combine into a single output \cite{chowdhary_natural_2020,gillioz_overview_2020,min_recent_2023}. 

Given these capabilities, NLP models have begun circulating among researchers and practitioners in the educational context. Research leveraging NLP is diverse, encompassing a range of analysis applications and utilizing its generative capacities \cite{sunar_natural_2024, bhaduri_nlp_2017, katz_utility_2023,shaik_review_2022}. These applications range from using NLP to extract features from text (e.g., \cite{berdanier_opportunities_2018, katz_exploring_2023, tai_examination_2024, verleger_using_2014}) to completing engineering assessments (e.g., \cite{gonzalez-calatayud_artificial_2021, nikolic_chatgpt_2023, savelka_can_2023}). When used to analyze textual data, the features, summaries, and categories produced using NLP can then be used for analysis. For instance, NLP enables the sentiment analysis of text, whereby the model extracts the underlying tonality of a piece of text and outputs whether it is positive, negative, or neutral \cite{crossley_modeling_2018, kastrati_sentiment_2021}. These analyses can then be used to gauge overall impressions and attitudes.

NLP models' text generation capabilities have also been used in education to design tutoring systems, educational content, and personalized learning experiences, among others \cite{mathew_nlp_based_2021,troussas_personalized_2023,madnani-etal-2018-writing}. These adaptive systems generate personalized recommendations and feedback for users, while adjusting the difficulty, level of support, and included material, among others. Instructors too benefit from NLP models which can automate the creation of educational content, saving time and resources while producing high-utility materials \cite{phillips_ai_2022,atapattu_comprehensive_2017,jayakodi_automatic_2015}.

Given the potential of NLP to analyze text, there is tremendous potential to leverage models for highly time- and resource-intensive processes, like qualitative thematic analysis. Thematic analysis is a widely used qualitative research method for identifying, analyzing, and reporting patterns or themes within data \cite{miles_qualitative_2014}. It is a flexible approach that can be applied to a range of data sources, including interview transcripts, focus group discussions, open-ended survey responses, and other forms of textual data \cite{miles_qualitative_2014}.

Many of the processes inherent to thematic analysis parallel the features of NLP \cite{katz_utility_2023, berdanier_opportunities_2018}. Consequently, investigations have used NLP for deductive code analysis \cite{tai_examination_2024, xiao_supporting_2023,katz2021using}, whereby the model is asked whether an existing code applies to a piece of text. However, inductive qualitative analysis, where the model produces a codebook without a reference, has been limited \cite{katz2021using}.

Our objective is to introduce a methodology for generating a thematic codebook utilizing NLP techniques, showcased through its application on a dataset comprising written feedback from engineering students at a research-focused university. In the next section, we detail these responses, known as student evaluations of teaching, and their significance in educational settings. This contextualization serves to lay the foundation for subsequent sections of our study.

\subsection{Student Evaluations of Teaching} 

Institutions have periodically facilitated SETs \cite{wachtel_student_1998, marsh_students_1987}. These surveys usually consist of Likert-scale items that are meant to provide students with the opportunity to ''rate" an instructor's teaching effectiveness, as well as open-ended prompts meant to elicit students' perspectives on their learning experiences, generating quantitative and qualitative data \cite{marsh_students_1987, sproule_student_2000}. Quantitative ratings, particularly those meant to measure overall teaching effectiveness, are commonly used by institutions as input into such administrative decisions as salary, promotion, and tenure \cite{hornstein_student_2017, abrami_dimensionality_2007}. Students, however, are more likely to engage in the evaluation process to share their perceptions about their learning experience in the class rather than how their ratings may influence administrative decisions, and they share these perspectives primarily through their responses to open-ended prompts \cite{hoel_why_2019}. These prompts generate qualitative data that is challenging to comprehensively distill for instructors teaching large class sizes, which results in the underutilization of these data. That underutilization results in a disconnect between how SETs are used in practice and students' motivation to engage in the process \cite{abrami_dimensionality_2007, hornstein_student_2017}.  

Natural language processing (NLP), a subset of machine learning, continues to grow in education research. NLP models can analyze and generate human language data, with applications like classifying text, generating content, and providing personalized learning experiences. SETs present one such use case for NLP, as the qualitative data from SET open-ended responses are currently underutilized. The following section describes the development of an NLP-based methodology for inductive codebook generation that can be used to analyze SET data more effectively, a process mirroring traditional inductive thematic coding.

\section{Methods}
\label{sec:methods}

In this paper, we are presenting a workflow that involves extracting information, embedding the extracted information in a high-dimensional vector space, clustering those embeddings, and summarizing the clusters. The goal of the extract, embed, cluster, and summarize (EECS) workflow presented here is to provide an easy-to-use and effective way to extract information from a large corpus of documents and distill it into a collection of themes. The process mimics the practice of codebook generation in traditional qualitative data analysis \cite{reyes2021living} but on a larger scale. In this case, those documents were SETs from introductory courses in the College of Engineering and College of Science at a large, research university. In such settings, it is common to have tens of thousands of SETs to analyze. The EECS workflow is designed to be scalable so it can be applied to a large corpus of documents beyond education settings. For demonstration purposes here we use a relatively small corpus of 5,000 SETs, but in theory, the workflow can be applied to a much larger corpus, depending on computing resources and the nature of the text units of analysis. Moreover, we believe the workflow scales beyond this specific context and can be applied to other types of documents, such as student essays, interview transcripts, and administrative records.

\begin{figure}[ht]
    \centering
    \includegraphics[width=0.75\textwidth]{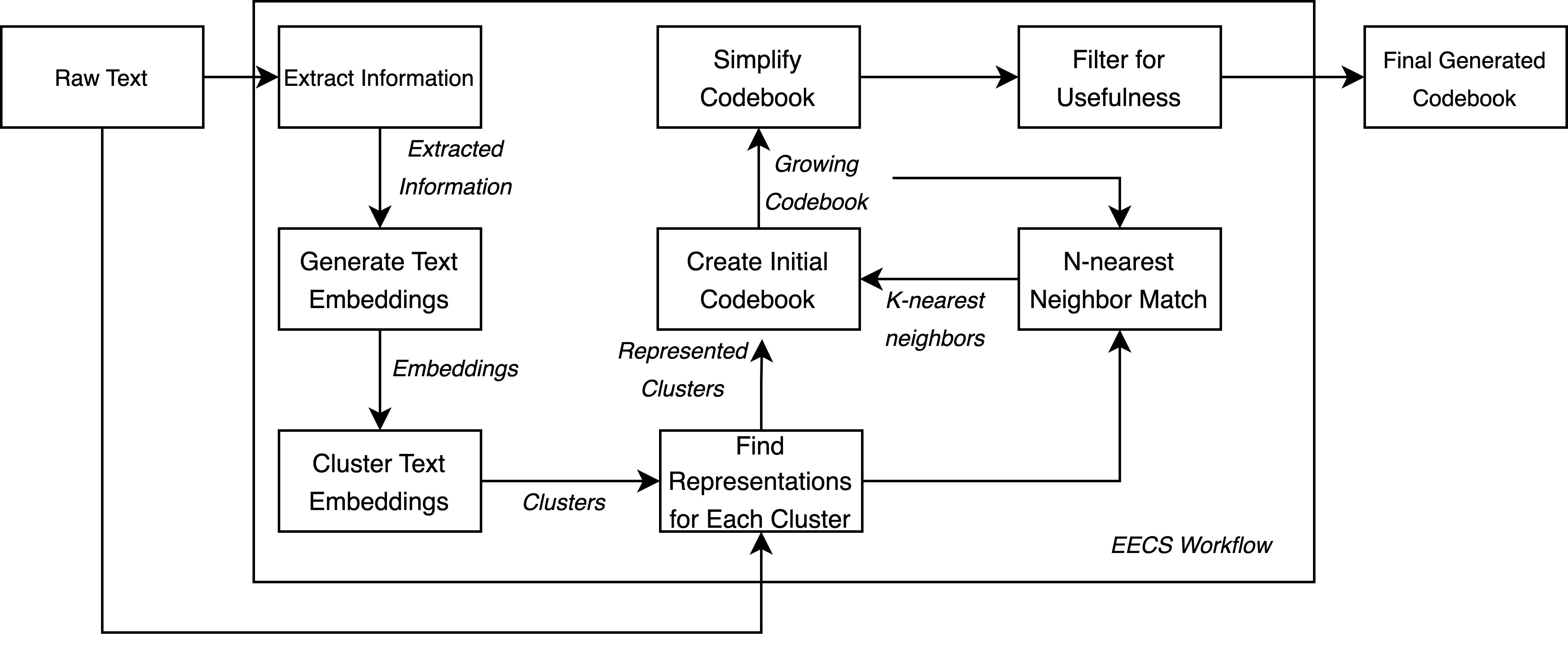}  
    \caption{Flowchart of the EECS Workflow}  
    \label{fig:eecs_workflow}
\end{figure}

Figure \ref{fig:eecs_workflow} shows a flowchart of the EECS workflow. The workflow begins with the extraction of ideas from the original documents. The extracted ideas are embedded into a vector space. The embedded ideas are then clustered into groups of similar ideas. Next, those clusters are summarized to create a codebook, which is simplified to remove redundancies and improve usefulness. The overall workflow is designed to be modular so that each step can be modified or replaced with a different technique or model as needed. Indeed, there is reason to believe that optimizing each step along the workflow would yield better results - a fruitful direction for future work. The workflow is also designed to be scalable so that it can be applied to large corpora of documents. Finally, we have designed this workflow to use open-source models and tools for the broader research community's use. These features allow researchers to run the EECS process on their own machine, thereby eliminating the need to send third-party providers, such as OpenAI or Google \cite{noauthor_privacy_2023,noauthor_gemini_nodate}. The following sections describe each step in the EECS workflow in more detail.

\subsection{Extracting Ideas from Original Texts}

The first step in the EECS workflow is to extract the information from the documents we want to analyze. This step is intended to overcome the varied language found in respondents' free-form SET responses because excessive variation could cause larger problems in the downstream analysis if too many similar options are present. This extraction step essentially is a standardization step and a way to isolate each of the ideas expressed in the SETs. In other words, we want to extract the main ideas from the SETs by disentangling them into simpler, discrete ideas. 

For this extraction step, we use a lightweight generative text model with the prompt given in the \nameref{sec:appendix}, section \ref{subsec:prompt_extraction}. The specific model was the dolphin fine-tuned version of Mistral-7b \cite{jiang2023mistral}. The extraction allows us to analyze each idea separately in subsequent iterations. For example, if a student said, ``I liked the in-class demonstrations and availability during office hours'', we would want to extract the two ideas ``in-class demonstrations'' and ``availability during office hours''. The prompt is designed to be a general prompt that can be used for any type of document so that the same model can be used for many other contexts. In practice, the model generates text from the prompt and original participant response, which is extracted in a short post-processing step.

\subsection{Embedding the Extracted Ideas}

The next step in the EECS workflow is to embed the extracted ideas into a vector space. By doing so, we aim to represent each idea as a vector, so ideas can be compared using mathematical operations such as cosine similarity, or be clustered together to find semantically similar ideas. We use a pre-trained text embedding model available from HuggingFace \cite{wolf-etal-2020-transformers} to embed the ideas into a vector space. The specific model used here was the UAE-Angle model \cite{li2023angle}, but other models would also work in this step. We used that model because it had the highest performance across a range of performance benchmarks at the time among models that could be run locally on one's machine. This was an important consideration to avoid sending data to third parties. 

The embedding model uses the text for each extracted idea as inputs and outputs a vector. The vectors from this UAE-Angle model are 1,024-dimensional, so the text for an idea is represented as a vector (i.e., collection of numbers) of length 1,024. The goal behind embedding models is to leverage the theory of distributional semantics, which asserts that words that appear in similar contexts have similar meanings \cite{erk2016you,boleda2020distributional}. Using distributive semantics, we can represent the semantic meaning of words and phrases as vectors. If two different phrases are semantically similar, as in ``used realistic examples'' and ``gave real-world scenarios'' then the vector representations of these should be close to each other in that high-dimensional vector space even though the two strings share no words in common. Working in this vector space allows us to compare the meaning of words and phrases using mathematical operations such as cosine similarity. Cosine similarity measures the angle between two vectors in a high-dimensional vector space \cite{kenter2015short}. If the angle between two vectors is small, then the cosine similarity is high, and if the angle between two vectors is large, then the cosine similarity is low. In text analysis, we can use cosine similarity to compare the meaning of two phrases. If the cosine similarity between two phrases is high, then the meaning of the two phrases is similar. If the cosine similarity between two phrases is low, then the meaning of the two phrases is different. This contrasts older text analysis methods, such as bag-of-words, which do not capture the semantic meaning of words and phrases but rely on the presence or absence of words \cite{khurana2023natural}. In that setting, ``used realistic examples'' and ``gave real-world scenarios'' would be treated as completely different phrases because they have different words, even though they are semantically similar.

In addition to the ability to calculate semantic similarity between pairs of texts, the embedding model also allows us to calculate the similarity between a single text and a set of texts, as one does when using clustering algorithms. Indeed, clustering is the next step in the EECS workflow because once we have the vector representations of the extracted ideas, we have a new problem: how do we find similar ideas so that we can reduce the information required to represent the SETs into something far smaller, i.e., on the order of 50-100 pieces of information. An alternative way to frame this problem is similar to principal component analysis and the idea of finding the dimensions that explain the most variance in a dataset. In this case, we want to find the dimensions that explain the most variance in the text of extracted ideas. Clustering can help us step in that direction by finding groups of similar ideas.

\subsection{Clustering Embedded Ideas Representations}

The next step in the EECS workflow is to cluster the extracted ideas. This is important because there are many instances of students commenting on similar ideas, e.g., ``great notes'' or ``the notes added online were very helpful''. To find the unique themes within the SETs, the extracted ideas must be collapsed into a more manageable number of observations. This can be accomplished using a clustering algorithm, which groups similar ideas. We used the clustering algorithm called hierarchical density-based spatial for applications with noise (HDBSCAN), which was implemented using the \lstinline{hdbscan} Python library \cite{mcinnes2017hdbscan}. While clustering, we used conservative parameter settings to encourage many small clusters. The goal was to have many clusters, each with a few ideas (i.e., ideally one). This was done to ensure that when we summarize the clusters later, we have a small number of ideas per cluster to summarize.  

\subsubsection{Finding Representative of Each Cluster}

With the clusters created, we now want to find a single representative member to pass to the next step in the workflow. For example, if we clustered 5,000 pieces of extracted information and formed 400 clusters, then we want 400 representatives - one for each cluster - to pass to the generative model. The goal is to have a more manageable quantity of information for the model to remember in its prompt. In essence, this step compresses 5,000 pieces of information to 400, assuming each cluster contains a large amount of redundant information. Finding a representative reduces this redundancy without losing information. 

To identify a representative, we first embedded a concatenated version of the ideas in each cluster. For example, if a cluster contained entries such as ``helpful office hours'', ``lots of office hours'', and ``good office hours'', we could combine them into a single string like ``helpful office hours lots of office hours good office hours'', that we can then embed. These cluster embeddings were then compared with the embeddings for the original extracted information to find the most similar piece of original extracted information. We used the same embedding model as in the prior embedding step and calculated the cosine similarity between the embeddings for each cluster and the embeddings for the original extracted information.  

Put another way, this representative identification step selects the extracted information with the highest cosine similarity to the cluster embedding. This ``most similar`` idea is assigned as the representative of that cluster. Let's imagine that the 295th cluster had grouped six pieces of extracted information. We concatenate those six ideas into a single string, which we then embed. We then calculate the cosine similarity between the embedding for that string and the embeddings for the extracted information from all the 5,000 original documents. Finally, we select the extracted information that had the highest cosine similarity with the cluster embedding and assign it as the cluster's representative. That representative is passed to the next step in the EECS workflow, summarization.

\subsection{Summarizing the Clusters}

Two more steps remain in the EECS workflow: creating the initial codebook (done in this summarization step) and simplifying it to remove redundancies and improve its utility. In the last step, we found representatives for each cluster that encapsulates its main themes. Now, the representatives are sequentially given to a generative text model with instructions to summarize it. The aim is to create an initial qualitative codebook with the most common themes in the SETs. This codebook can then be simplified and used in downstream analyses to understand the prevalence of the themes in the SETs. The codebook application step is beyond the scope of this paper and will be described in future work. However, to generate the codebook, we can use the prompt given in the \nameref{sec:appendix}, section \ref{subsec:prompt_summarizing}. 

This initial generative step is part of a retrieval-augmented generation (RAG) pipeline. One challenge in generating a codebook is that excessive entries are found, with many being redundant. Consequently, the prompt included instructions for the model to reference an existing codebook (i.e., the one we are inductively creating), and decide whether the extant codes cover the theme(s) in the new text under analysis. If not, a new code entry is warranted because the existing codes are insufficient. 

While it might seem logical to provide the entire codebook to the model as part of the prompt, our practical experience has shown that it's more effective to offer the top $k$ codes from the codebook. By ``top $k$ codes," we refer to the $k$ most probable codes based on semantic similarity between the cluster text and existing codes. This approach reduces the memory requirements on the model; rather than memorizing the entire codebook, the model only needs to recall the top $k$ codes. This is a tradeoff because the model may not always make an informed decision about the necessity of a new code if it only has the top $k$ codes. However, in our experiments, the model demonstrated the ability to make such informed decisions, as evidenced by the reasoning steps it provided in the output.

To determine the top $k$ codes, we embed each entry in the codebook using the same embedding model as in the prior steps and calculate the cosine similarity between the embeddings for each codebook entry and the embeddings for the cluster representative. Subsequently, we select the top $k$ entries in the codebook with the highest cosine similarity with the cluster representative embedding as the top $k$ codes. For instance, if the new cluster representative is ``gave lots of examples in class," the top $k$ codes might include ``in-class demonstrations," ``real-life scenarios," and ``helpful lectures." These codes are then included in the prompt provided to the model.

In response to the prompt (detailed in the \nameref{sec:appendix}, section \ref{subsec:prompt_summarizing}), the model generates a summary of the cluster with a code, definition, and example if it determines that a new entry in the codebook is necessary. We subsequently extract the generated text in a short post-processing step. Ultimately, the text generated after the prompt serves as the codebook entry, which is utilized in the subsequent simplification step. Despite its complexity, the process is conceptually straightforward: create a codebook by examining a representative sample of extracted ideas, while striving to keep it concise by adding new entries only when essential. This necessity is determined by evaluating the existing content in the codebook. In essence, the EECS workflow mirrors traditional qualitative data analysis practices.

\subsection{Simplifying the Codebook}

The concluding step in the EECS workflow involves the simplification of the codebook. This step is crucial as the initial codebook generated can often be lengthy and redundant, potentially complicating its utility in subsequent analyses. The aim is to streamline the codebook for easier application in downstream processes. 

We employ a generative text model utilizing a prompt outlined in the \nameref{sec:appendix}, as detailed in section \ref{subsec:prompt_simplifying}. The design of this prompt ensures its adaptability for diverse document types, enabling the use of the same generative text model across various contexts. In practice, the model generates text based on the provided prompt and the original codebook. Subsequently, we extract the generated text, constituting the final version of the inductively constructed codebook. This approach ensures that the codebook is distilled to its essence, facilitating its practical application in subsequent analyses.

\section{Results}

\subsection{Extracting Ideas from Original Texts}

\begin{figure}[ht]
    \centering
    \includegraphics[width=0.75\textwidth]{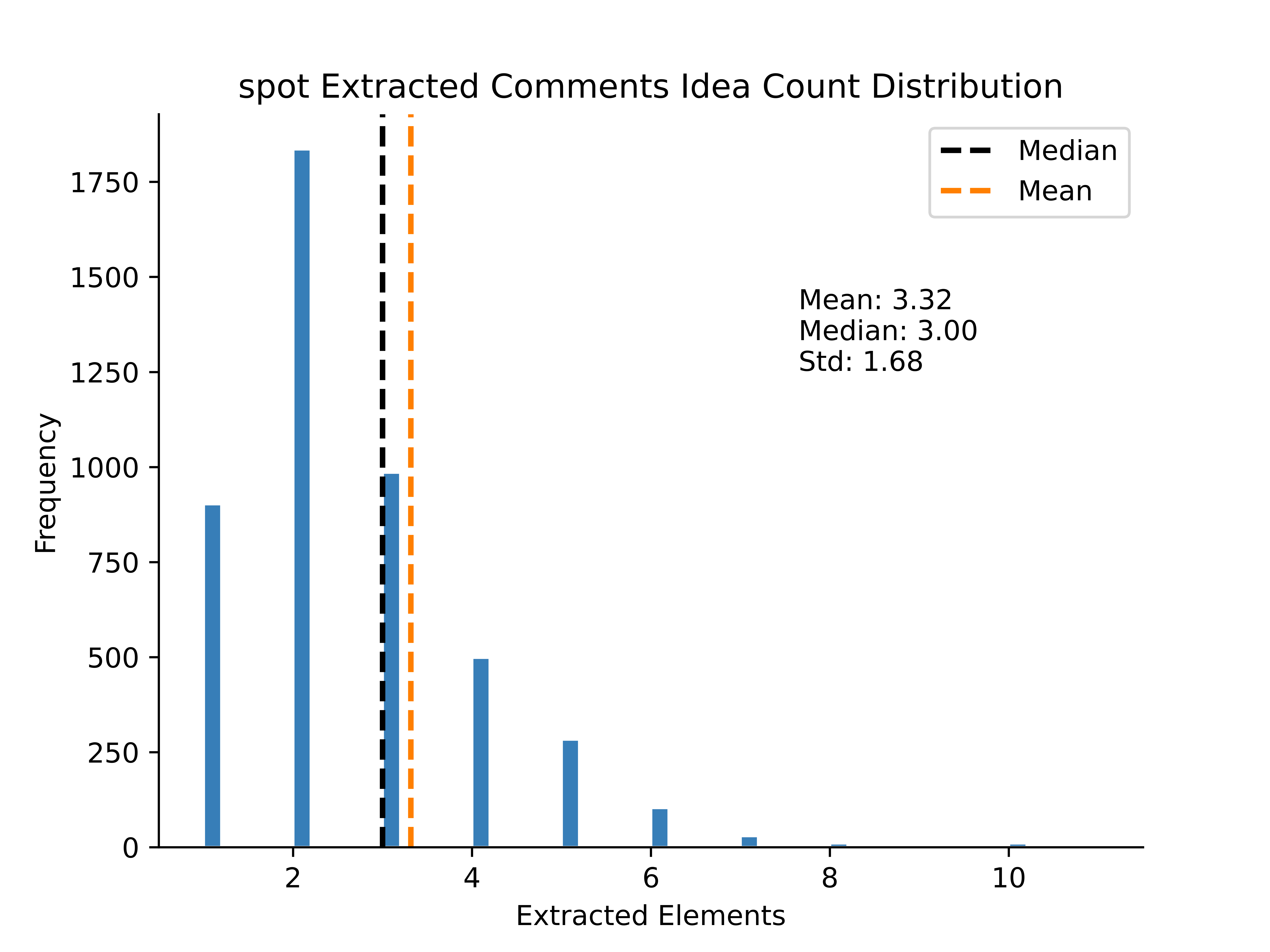}  
    \caption{Histogram of Number of Extracted Ideas from Original Comments}  
    \label{fig:extracted_ideas}  
\end{figure}

After pre-processing and screening short responses from our random sample of 5,000 responses, we ended up with 4,672 unique original comments. Using the EECS workflow, we first extracted 12,046 ideas from these 4,672 original SETs. The distribution of the extracted ideas is shown in Figure \ref{fig:extracted_ideas}. The average number of extracted ideas was 3.3 with a median value of 3. The right skew of the distribution was expected because most students only have a few things to say while a few outliers have much more that they want to share.

Table \ref{tab:extracted_ideas} shows an example of the extracted ideas from a single SET. In this example, two ideas were extracted: the instructor provided examples and the instructor explained the problem-solving process.

\begin{table}[ht]
    \caption{Extracted ideas from a single SET}
    \label{tab:extracted_ideas}
    \centering
    \begin{tabularx}{\textwidth}{XX}
    \toprule
    \textbf{Student's Text} & \textbf{Model Response} \\
    \midrule
    She did examples and explained how problems were done. & My summary: \newline 1. Provided examples \newline 2. Explained problem-solving process \\
    \bottomrule
    \end{tabularx}
\end{table}

\subsection{Summarizing the Clusters}

We then clustered these 12,046 extracted pieces of information into 272 clusters. Summarizing the 272 clusters yielded a codebook with 232 codes. An example of the input and output for this step is given in Table \ref{tab:sum_cluster}. Note that this step ran with two slightly different prompts. For the first $k$ steps, the process ran without a RAG implementation since there were not enough codes in the codebook for the RAG part of the prompt. After $k$ steps, we then used the prompt in the \nameref{sec:appendix}, section \ref{subsec:prompt_summarizing}, which included the top $k$ codes from the codebook as the RAG implementation. Doing this split of the prompts ensured there were enough entries in the codebook to start using for the nearest-neighbor matching part of the RAG process. 

\begin{table}[ht]
    \caption{Example of summarizing clusters}
    \centering
    \begin{tabularx}{\textwidth}{>{\raggedright\arraybackslash}X>{\raggedright\arraybackslash}X}
    \toprule
    \textbf{Prompt} & \textbf{Model Response} \\
    \midrule
    \textbf{Input} & \textbf{Output} \\
    the instructor used powerpoint presentations effectively & Label: Effective PowerPoint Use \newline Definition: The use of well-designed and engaging PowerPoint slides by instructors. \newline Example: ``The instructor used powerpoint presentations effectively.'' \\
    \bottomrule
    \end{tabularx}
    \label{tab:sum_cluster}
\end{table}

\subsection{Simplifying the Codebook}

From the 232 codes generated in the previous step, we simplified the codebook to 159 codes. This simplification proceeded in two steps. The first was a RAG implementation that asked the model to check each code and for a similar one already in the codebook. An example of that output is given in Table \ref{tab:simp_cb_step}.  

\begin{table}[ht]
    \centering
    \caption{Example of Step 1}
    \begin{tabularx}{\textwidth}{>{\raggedright\arraybackslash}X>{\raggedright\arraybackslash}X}
    \toprule
    \textbf{Input} & \textbf{Output} \\
    \midrule
    quick email response & Step 1: Examine existing codebook.
    [`insufficient detail', `well prepared organization', `lack of action suggestions', `feedback quality', `response speed'] \newline Step 2: Read new written response in \textless text\textgreater tag. ``quick email response time''. Step 3: Determine if there is an existing label to describe the text. The given text discusses ``email response time,'' which seems related to ``response speed.'' Therefore, I will use this code from the existing codebook.    Step 4: If no suitable label exists in the existing codebook, generate a new one and provide its definition and an example from the \textless text\textgreater tag. Since there is already a relevant label (`response speed'), no new label needs to be suggested. No new label suggested. \\
    \bottomrule
    \end{tabularx}
    \label{tab:simp_cb_step}
\end{table}

The second step in the codebook simplification process was a check on whether the code was sufficiently clear. An example of the input and output for this step is given in Table \ref{tab:simplified_codebook}. In the example, the original code ``encouraging active participation" was reviewed and the model suggested keeping the code because it was specific and clear enough for its intended purpose.

\begin{table}[ht]
    \centering
    \caption{Example of simplification step with accepted original code}
    \begin{tabularx}{\textwidth}{>{\raggedright\arraybackslash}X>{\raggedright\arraybackslash}X}
    \toprule
    \textbf{Input (Code)} & \textbf{Output (Judgment)} \\
    \midrule
    encouraging active participation & My response: The given code, `encouraging active participation', seems like a good code as it specifically addresses an aspect of teaching that instructors might be interested in receiving feedback on from their students. It is clear and informative for the research team to understand what kind of feedback they are looking at when analyzing responses related to this theme.

    My suggestion: Keep the code.
    
    Alternate suggestion: None, as the current code is already specific and clear enough for its intended purpose.
    \\
    \bottomrule
    \end{tabularx}
    \label{tab:simplified_codebook}
    \end{table}

An example of an alternative suggestion from the model is given in Table \ref{tab:simplified_codebook_alt_sug}. In this example, the model was given the code ``clarification and simplification'' and suggested discarding this code and creating two new codes - ``clarification'' and ``simplification'' - to better understand how students responded to feedback forms.

\begin{table}[!h]
    \centering
    \caption{Example of simplification step with an alternative suggestion}
    \begin{tabularx}{\textwidth}{>{\raggedright\arraybackslash}X>{\raggedright\arraybackslash}X}
    \toprule
    \textbf{Input (Code)} & \textbf{Output (Judgment)} \\
    \midrule
    clarification and simplification & My response: The given code, `clarification and simplification', seems to be addressing two distinct aspects related to feedback forms - making instructions or questions clearer (clarification) and presenting them in a more straightforward manner (simplification). While both are important for improving the quality of responses from students, they can also be considered as separate codes. This would allow researchers to better understand which aspect is being addressed by each theme.

    My suggestion: Discard this code and create two new codes - `clarification' and `simplification'. By separating these themes into distinct codes, the research team will have a more precise understanding of how students are responding to feedback forms. This can help identify specific areas for improvement in future iterations of the form design process.
    
    Alternate suggestion:
    - Clarification: Refers to making instructions or questions on feedback forms clearer and easier to understand.
    - Simplification: Addresses presenting information on feedback forms in a more straightforward manner, reducing complexity and potential confusion for students.
    \\
    \bottomrule
    \end{tabularx}
    \label{tab:simplified_codebook_alt_sug}
    \end{table}

\subsection{Comparison with the Human-Generated Codebook}

As previously mentioned, a subset of SET data was originally employed in prior studies investigating the teaching and learning dynamics within first-year engineering courses \cite{soledad_leveraging_2017, soledad_understanding_2019}. These SETs underwent manual thematic coding, revealing 39 distinct codes \cite{soledad_understanding_2019}. 

Established frameworks provide one way of synthesizing responses to open-ended responses in SETs \cite{creswell_research_2014}. Soledad \cite{soledad_leveraging_2017} leveraged the Academic Plan Model \cite{lattuca_shaping_2009} for this purpose, which acknowledges the crucial role of students' perspectives in curriculum planning. APM has informed prior work investigating students' learning experiences and instructors' curricular decision-making processes (e.g., \cite{knight_influence_2016, knight_exploration_2012}). It provides a structure for categorizing and presenting qualitative data on students' learning experiences and their perspectives on the learning environment. According to the APM, the learning environment consists of eight elements, described in Table \ref{tab:apm_element_descriptions}.

The initial coding revealed that the analyzed SETs corresponded with these APM elements \cite{soledad_leveraging_2017}. However, the 39 human-generated codes fell into only four categories: one in \textit{Adjustments}, one in \textit{Assessment}, 13 in \textit{Faculty characteristics}, 18 in \textit{Instructional processes}, and six in \textit{Instructional resources}.

In contrast, our EECS method generated 80 distinct codes. Five were excluded due to ambiguity or vagueness, yielding 75 usable codes, of which six pertained to \textit{Adjustments}, five to \textit{Assessments}, 22 to \textit{Faculty characteristics}, 25 to \textit{Instructional processes}, 15 to \textit{Instructional resources}, and two to \textit{Content and Sequence} — a category not initially identified through manual coding. \textit{Content and Sequence} refers to the arrangement and content of subject matter in a course. This difference between the human and the EECS process may indicate that the larger number of SETs processed contain information beyond the smaller initial sample used for the human coding - a promising finding for similarly large datasets. The complete list of EECS-generated codes alongside the human-generated codes is provided in the \nameref{sec:appendix}, section \ref{subsec:code_comparison}. 

Comparing the model's codes with the APM demonstrates the method's alignment with traditional human-generated codes. Because the human-generated codebook aligned with the APM elements, we should expect the model's codes to align. Once categorized according to the APM, the human- and model-generated codes are more meaningful side-by-side, allowing for a clearer comparison. 

Examining the distribution of codes across categories, our method mirrors a similar trend to the manual analysis. Notably, the model's codes were categorized independently of the original codebook and were cross-checked by the researcher who conducted the initial coding. Furthermore, there is substantial overlap between the original codes and model-generated codes within each category, indicating that the model effectively captures analogous themes, terms, and characteristics. Following the final simplification step, the model's codes were largely constructive and informative, facilitating their integration into subsequent analyses or feedback for instructors.

Moreover, the significant concordance between the method's codes and those manually derived indicates the model's proficiency in encapsulating the essence of the analyzed SETs. This finding holds promising implications for the research community, as it suggests that inductive thematic analyses can be efficiently conducted on extensive datasets using NLP within a relatively short timeframe. These implications will be further explored in the \nameref{sec:discussion}.

\begin{table}[!h]
    \centering
    \caption{Elements of the Academic Plan Model adapted from \cite{soledad_understanding_2019}}
    \begin{tabularx}{\textwidth}{>{\raggedright\arraybackslash}X>{\raggedright\arraybackslash}X}
    \toprule
    \textbf{Element} & \textbf{Description} \\
    \toprule
    Adjustments & Enhancements and adjustments to the academic plan based on experience, evaluations, and student feedback \\
    \midrule
    Assessment & The strategies used to determine whether decisions about elements of the academic plan are optimal \\
    \midrule
    Faculty Characteristics & Perceptions about the availability, approachability, and focus of instructors \\
    \midrule
    Instructional processes & The instructional activities by which learning may be achieved, such as scaffolding, assessment practices, and in-class problem descriptions \\
    \midrule
    Instructional resources & The materials, technologies settings used in the learning process, and the opportunities to engage with the instructor outside of class time \\
    \midrule
    Content \& Sequence & The arrangement and content of subject matter toward achieving learning outcomes \\
    \bottomrule
    \end{tabularx}
    \label{tab:apm_element_descriptions}
\end{table}

\section{Discussion}
\label{sec:discussion}

\subsection{Comparison to Traditional Codebook Generation}
These results highlight several important points. First, the EECS workflow can replicate the process of codebook generation in a traditional qualitative data analysis setting. Not only did the process reproduce codes from traditional analysis but also it identified new codes with more granularity than the original codebook. These new codes may be due to the model's ability to identify more subtle differences in the data than a human coder. Alternatively, this may be because we can simply analyze more data. Whereas the original analyses only looked at a few hundred SETs, the EECS workflow was able to analyze over 4,500 unique SETs. This is a significant increase in the amount of data that can be analyzed and may have led to the identification of new codes because something that barely showed up in the original analyses was now able to be detected when there were more data to consider. 

Second, as mentioned in the \nameref{sec:methods} section, the EECS workflow is designed to be scalable. This is important because the process can be applied to a large corpus of documents. In this case, we used a relatively small corpus of a few thousand SETs, but in theory, the workflow can be applied to a corpus much larger, depending on computing resources and the nature of the text units of analysis. One area for future exploration is the extent to which the workflow can be used with longer texts in less constrained settings. For example, the workflow could be used to analyze student essays, research articles, and administrative records. Compared to those alternative data types, SETs are relatively short and constrained, so it is an open question whether the workflow would be as effective in those other settings.

\subsection{Comparison to Other NLP Methods}

While the EECS workflow was able to replicate and extend human coding efforts, it is important to consider the workflow's performance in comparison with other popular approaches to computer-assisted thematic analysis. One popular method for analyzing text data is topic modeling. Topic models are statistical models used to identify topics in a collection of documents. One of the most popular topic modeling algorithms is latent Dirichlet allocation (LDA) \cite{blei2003latent}. The LDA model is a generative probabilistic model that assumes that each document is a mixture of topics and that each word in the document is attributable to one of the document's topics. The goal of an LDA model is similar to the goal in the EECS workflow: identify the latent topics that are giving rise to the observed words in each unit of analysis (e.g., document, essay, article, SET). In practice, an LDA model is trained on a corpus of documents and then used to identify the topics in the documents. One advantage of an LDA model is its quick training - LDA models can be trained in a matter of minutes. In contrast, the EECS workflow can take hours to run depending on corpus size and computation resources. Another comparative advantage of topic models is that there are specific metrics one can use to determine the number of topics to extract from the data \cite{wallach2009evaluation}. The EECS workflow does not have such a metric at the moment and instead relies on the user to determine the number of clusters to extract from the data.

On the other hand, the EECS workflow has several advantages. One big advantage is that the EECS method relies more on semantics than syntax. Typical topic models rely on word stems so words like ``homework'' and ``assignments'' are treated differently even though one may want to consider them similar. More modern variations of topic models such as BERTopic \cite{grootendorst2022bertopic} attempt to address this limitation of traditional topic models, to limited degrees of success. Another advantage of the EECS workflow is the generation of coherent labels. Often, in traditional topic model implementations, a researcher is given a list of the most frequent words in each topic and then left to determine what that topic might be about. Such divination can yield fuzziness in the eventual results. In contrast, the EECS workflow can not only generate coherent labels but also accompanying definitions.

BERTTopic harnesses the power of transformer-based models, the cutting-edge artificial neural network architecture that has revolutionized many NLP tasks \cite{gillioz_overview_2020}. Introduced by Vaswani et al. in their seminal paper \cite{vaswani2017attention}, transformers have rapidly become the foundation of state-of-the-art NLP models due to their remarkable ability to capture long-range dependencies in sequential data like sentences \cite{gillioz_overview_2020}. By weighing the importance of different words in a sequence and processing data in parallel, transformers can learn intricate patterns in language, making them invaluable for NLP applications \cite{gillioz_overview_2020, katz_utility_2023}. 

Our method, like BERTTopic, employs transformer-based models for the codebook generation process. However, we are not unique in recognizing the potential of transformers for NLP-based qualitative thematic analysis. Ganesh et al. \cite{ganesh_response_2022} introduced the response construct tagging (RCT) task to analyze students' open-ended responses from course surveys, focusing on identifying affective states related to transformative experiences and engineering identity development. Their work showcased the scalability and value of transformer models like RoBERTa \cite{liu_roberta_2019} for qualitative coding in educational research contexts.

Similarly, Rahman et al. \cite{rahman_analyzing_2021} used transformer-based NLP to extract the required competencies for software testing professionals from online job postings. By comparing these themes to course syllabi from 20 universities, they found gaps between employer-valued skills and software testing curricula. This research highlights NLP's scalability and utility for analyzing unstructured data and its synergy with traditional thematic analysis techniques.

Beyond theme extraction, Wang et al. \cite{rose_e_wang_sight_2023} explored using ChatGPT to classify YouTube comments on the MIT OpenCourseWare (OCW) channel. The high correlation between human inter-rater reliability and human-model agreement across categories suggests that LLMs mirror human judgment variability in annotation tasks. This work underscores the insights derived by combining NLP with traditional qualitative methods for large, unstructured text data in education \cite{rahman_analyzing_2021}.

While these investigations demonstrated NLP's supportive role in traditional thematic analysis, Tai et al. \cite{tai_examination_2024} proposed a novel methodology using LLMs for deductive coding in qualitative research. Parallel to our research group's work \cite{gamieldien_utilizing_2023, katz2021using}, they developed a deductive coding scheme with five predefined codes related to scientific identity and input sample interview excerpts and the codebook into ChatGPT to detect code presence and supporting evidence. Comparing the results to human coding, the authors found high alignment for most codes. Interestingly, areas of divergence prompted refinements in code definitions and applications, showcasing how LLMs can serve as external reviewers to identify potential blind spots, a finding we also observed. Transformer-based NLP techniques have shown tremendous potential in assisting and guiding qualitative thematic analysis, but research thus far has been limited to a few approaches and has not ventured into inductive thematic analysis as our method proposes.

Accordingly, there are several potential reasons our inductive process works. To start, the initial information extraction step helps to standardize language and disentangle complex ideas into simpler, singular ideas. This is important because it allows us to analyze each idea separately in subsequent steps. Without doing this, we would have to analyze the entire SETs at once, which would be a much more difficult task. Additionally, the embedding step further helps funnel information into standardized formats because it captures the semantic meaning of words and phrases. Therefore, we have a way to move from the varied original text into an abstract space of ideas. Prior methods using dictionary-based approaches simply were not able to do this. Next, the dimension reduction and clustering steps help to reduce the information even more than the prior steps. The result is to have a funnel that takes in a large amount of varied information and outputs a small number of themes.   

\subsection{Humans-in-the-Loop}


While our methodology is largely automated and programmatic, human involvement is indispensable. Human expertise and judgment play a pivotal role throughout the process, ensuring the credibility of the analysis and validating the outputs. For instance, employing the SETs, our method generated 80 total codes, but five were discarded due to ambiguity or irrelevance. One such code, ``lack of specific feedback," emerged from instances where students either provided ``No comment" or left the SET item blank. The EECS process yields a codebook summarizing the themes and content of the input data, yet it may include themes and content that are invalid, incoherent, or misaligned with research objectives. We advocate for researchers to actively engage in the codebook generation process, identifying irrelevant or misleading codes, manually reviewing and rejecting unnecessary ones, and verifying the overall quality and validity of the automated analysis.

Code quality also pertains to its usefulness. While the EECS process generates a codebook, the workflow by default is agnostic about how codes will be utilized - a result of embedding, clustering, and prompt designs. Hence, researchers must assess the relevance and applicability of generated codes to research questions, considering the required level of granularity and specificity for meaningful analysis. Establishing criteria for determining the saturation point in code generation is advised, as this varies between investigations. Researchers must monitor the codebook to identify when no new themes or insights emerge from the data. In our case, we observed that the model began outputting overly vague codes, signaling the end of the generation process. Balancing code granularity and manageability is crucial, as codes become simpler with each iteration. Consideration for potential consolidation or hierarchical organization during code review can help conclude the generation process while maintaining an ideal level of granularity.

The prompts used in the EECS process are detailed in the \nameref{sec:appendix} and are specifically tailored for this application. If applying this process to a different context, crafting context-specific prompts is essential to guide the NLP model effectively. Prompt design is iterative and should reflect the quality and relevance of generated codes. Balancing specificity and flexibility in prompt design is crucial to avoid scope limitations or researcher biases \cite{maoBias, sheng-etal-2019-woman}. Additionally, there is potential for this method to be adapted to fine-tuned LLMs trained on domain-specific datasets, potentially enhancing accuracy, relevance, and efficiency \cite{Church_Chen_Ma_2021}. However, fine-tuning poses challenges such as computational resources, model interpretability, and ethical considerations, which should also be considered.

The bottom line is this: the EECS workflow works in general. Yet, to maximize utility, researchers should actively participate in the codebook generation process, from assessing code quality and utility to designing prompts and selecting models. This method is inherently reflective, requiring researchers to play an active role in code generation and analysis rather than passively relying on method outputs.

\subsection{Limitations}
The current workflow presents several limitations, some of which suggest areas for future exploration. Firstly, granularity levels within the codebook pose a challenge. Instances such as having separate codes for ``in-class demonstrations" and ``used experiments to show ideas" may exhibit semantic similarity, suggesting potential consolidation into a single code. Similarly, codes like ``clarity" and ``clarity on syllabus" could be collapsed or treated hierarchically, especially if clarity is evident in other aspects of the course. Addressing this challenge is a focus for future work, necessitating the development of methods to programmatically handle semantically similar codes, including refining techniques for consolidation and hierarchical treatment.

Secondly, the efficacy of codebook generation heavily relies on the models employed in the process. As models evolve, ongoing evaluation of the method's efficacy and accuracy, particularly with diverse datasets, is essential. Variations in model size and fine-tuning techniques can significantly impact outcomes, requiring thorough evaluation across different models and datasets to ensure robustness and reliability.

Thirdly, the interpretation of codes exhibits context dependency. For instance, the interpretation of ``scaffolding" may vary when analyzing SET data compared to comments from a construction site. Exploring methods to incorporate context into the codebook generation process, whether through prompt adjustments or alternative approaches, is an avenue for future investigation.

Finally, while the process mimics human coding efforts, its superiority over alternative methodologies remains uncertain. Alternative approaches may exist, utilizing different prompts, variations in the RAG pipeline implementation, or alternative methods of codebook simplification. Moreover, advancements in model architectures, such as selective state space models, may mitigate context window limitations and introduce novel possibilities. While our work demonstrates the potential of using generative text models and modern NLP techniques with open-source options to replicate human coding processes efficiently and at scale, there are diverse avenues for exploration in designing processes for document analysis. We encourage further research to build upon this foundation, facilitating a range of options for document analysis across various fields.

\section{Conclusion}

Recent advances in NLP and LLMs promise opportunities to analyze text data in new ways. In this paper, we have discussed a new analytic method for such analysis. By combining information extraction, text embedding, clustering, and summarizing, we have shown that it is possible to identify themes in a large corpus of documents without the typical time and labor costs associated with manual coding. The process we have described is scalable and can be applied to create a codebook that can be used in downstream analyses. We believe that this method has the potential to be useful for analyzing SETs and other types of student writing in teaching and research settings. We also believe that this method has the potential to be useful for analyzing other types of documents, such as student essays, research articles, and administrative records. We hope that this paper will inspire others to explore the possibilities of NLP and LLMs for analyzing documents in their fields. 

\section{Acknowledgements}
We thank the reviewers for their helpful comments and suggestions. We also thank the students who participated in the study. This work was supported by the National Science Foundation under Grant No. 2107008.


\section{Appendix}
\label{sec:appendix}

\subsection{Prompt for Information Extraction}
\label{subsec:prompt_extraction}

\begin{quote}
You are an expert text analyst reading {data type}s collected in \{data collection context\}. I am going to send you one of these \{data type\}s. I need you to use your expertise to analyze the provided text in the \textless text\textgreater tag below and summarize it in an enumerated list. You should do this analysis by providing several short descriptive phrases that summarize each idea discussed in the \{data type\} that answered the prompt. When you suggest multiple items, separate each one in your response with a new line. You MUST remove anyone's names and *use gender neutral pronouns* for deidentification purposes. Start your response with ``My summary:''. Here is an example of input and desired output from a different context when there are only two topics, but remember that you can suggest as many topics as you think are necessary for the text you summarize.
Example input: ``Jared did a great job responding quickly to emails and turning in good work.''
Example output: ``My summary: 
1. Responded quickly to emails 
2. Turned in good work''.
Notice how the main ideas are summarized and there are no names or pronouns included here. Also, notice how the response did not make up information that was not in the input. You must NEVER make up information that is not in the input text you receive because there is a severe penalty for that. If the text you receive is very short and says ``nothing'', do not make up new things.
Here is the text for you to summarize: \textless text\textgreater \{text\} \textless /text\textgreater 
Begin your analysis now.
\end{quote}

\subsection{Prompt for Summarizing Clusters}
\label{subsec:prompt_summarizing}

\begin{quote}
Act as if you are \{persona\}.
You specialize in creating short labels to summarize the main idea being discussed in a collection of text.
The labels you generate will be used to systematically label all the \{data type\}s to identify key themes in an entire dataset of text. 
I need your help analyzing a cluster of \{data type\}s from \{data collection context\}.
In theory, the cluster of {data type}s should be semantically similar and related to each other, but occasionally it may not be homogenous.
Your task is to generate a small number of labels that capture the main concepts or themes discussed in the majority of the {data type}s you review.
To do this, please start by reviewing the {data type}s provided to you in the \textless text\textgreater tag. 
Each of the {data type}s is separated by a new line.
After reading the {data type}s in the cluster, please then generate your summary labels. 
For each label you generate, please provide:
- The label (a short phrase);
- A brief definition of what the label represents.
- An example of a {data type} from the actual text in the \textless text\textgreater tag.
The goal is to capture the essence and meaning of the *majority* of comments in each cluster of {data type}s through the labels applied. 
Your response should start 'Label: ' followed by your label and then on a new line 'Definition: ' followed by your definition for that label.
For example: 
Label: \textless label 1\textgreater 
Definition: \textless definition 1\textgreater
Example: \textless example 1\textgreater
As a reminder, here is the cluster of {data type}s for you to analyze:
\end{quote}

\subsection{Prompt for Editing the Codebook}
\label{subsec:codebook_editing}

\begin{quote}
Act as if you are the world's best qualitative data analyst with expertise in generating qualitative codebooks for thematic analysis.
Your task is to do the following four steps. Step 1: examine an existing codebook given to you in the \textless existing codebook\textgreater tag. Step 2: read a new code given to you in the \textless text\textgreater tag. Step 3: determine whether the existing codebook already has enough labels to describe the possible new code in the \textless text\textgreater tag. Step 4: Either accept the new code because it is sufficiently different from other codes or reject the new code because it is redundant. This is admittedly subjective, but you should be able to use your expert judgment to decide whether or not existing codes together provide sufficient coverage at an abstract level. Either way, be sure to provide your reasoning for whether or not you think you need to accept or reject the possible new code. Please note that there is a penalty for adding redundant codes, so you should only create a new code if you are certain the existing ones are insufficient. You should start your response with ``My analysis:'' followed by your step-by-step reasoning. You should conclude your response with ``My final verdict:'' followed by ``Keep code \textless code\textgreater'' or ``Reject code \textless code\textgreater'' where \textless code\textgreater is a placeholder for the actual label you are judging. 
For your reference, here is the existing codebook \textless existing codebook \textgreater \{codes\} \textless /existing codebook\textgreater.
\end{quote}

As before, note that ``\{codes\}'' is a placeholder where the top $k$ codes are placed for each With this prompt, the model then generates a summary of the cluster with a code, definition, and example. 
    
\subsection{Prompt for Simplifying the Codebook}
\label{subsec:prompt_simplifying}

\begin{quote}
    Act as if you are the world's best social science researcher. You specialize in qualitative data analysis and thematic coding. I need your help. I have a list of codes in a codebook to label themes in \{data type\}s collected from a \{data collection context\}. These codes need to be useful and informative for the research team, meaning they need to be specific and clear. Your task is to read a code from the codebook that I will give you in the \textless text\textgreater tag and tell me whether or not the code is a good code that I should keep in the codebook or a bad code that I should discard. Format your response as ``My response:'' followed by your step-by-step reasoning about whether it is a good code. Then, on a new line, say ``My suggestion:'' followed by your suggestion for whether or not to keep the code. If you think we should discard the code then you should add yet another line that says ``Alternate suggestion:'' followed by a better version of the code you reviewed. Here is the code for you to review: \textless text\textgreater zoom meeting requirement \textless /text\textgreater
\end{quote}

\newgeometry{margin=1cm, includefoot}
\begin{landscape}
\subsection{EECS vs. Manual Code Comparison}
\label{subsec:code_comparison}
\begin{table}[!htb]
\centering
\resizebox{\textwidth}{!}{\begin{tabular}{llll}
\toprule
Adjustments & Adjustments (EECS) & Assessment & Assessments (EECS) \\
\midrule
Student feedback-based adjustments & improved understanding & Emphasis on conceptual understanding & comprehensive resource assessment \\
& course improvement recommendations & & content understanding \\
& student experience and satisfaction & & conceptual difficulties \\
& course feedback and dissatisfaction & & comprehensive content coverage \\
& positive feedback & & time constraints and preparation issues \\
& feedback quality (timeliness \& satisfaction) & & \\
\bottomrule
\end{tabular}}
\caption{EECS vs. Manual Code Comparison (Part 1)}
\end{table}

\begin{table}[!htb]
\centering
\resizebox{\textwidth}{!}{\begin{tabular}{llll}
\toprule
Faculty Characteristics & Faculty Characteristics (EECS) & Instructional Processes & Instructional Processes (EECS) \\
\midrule
Being accessible & teaching quality and effectiveness & ``Behavior, classroom engagement" & engineering problem-solving strategies \\
``best professor" & motivational factors & Online utilities & engaging presentation style and note taking \\
``organized" & instructor support, responsiveness, and appreciation & Work-out problems and examples & effective use of visual aids and note-taking effectiveness \\
Time management & clarity & ``Presented material very clearly" & task preparation \\
``well-prepared" & simplification & Assessment practices & detailed explanation \\
``Behavior, Classroom engagement" & instructor-student interactions & Clarify prior misconceptions & mathematical explanation \\

Being accessible & classroom environment management & Demonstrate structures and processes to solve work-out problems & student engagement \\
Emphasize and discuss core concepts & support for struggling students & Emphasize and discuss core concepts & classroom interaction \& participation \\
Foster student success & support and assistance & Homework & active learning strategies \\

General perception of teaching strategy & positive impact and accessibility & Instructor-provided notes & lab experiences \& adoption \\
Provide guidance and feedback & flexibility in scheduling & Multimedia & supervision \& safety procedures \\
Show concern and interest in students' lives & consistency in schedule & Office hours & structured lessons \& teaching methods \\
& public humiliation & Practical applications & practical application and feasibility \\

& inclusive atmosphere & Provide guidance and feedback & value of examples and demonstration \\
& effective communication & Real-world applications & real-life applicability and relevance \\
& clear communication & Review sessions & technical help \\
& communication strategies and styles & Scaffolding & skill improvement \\
& focused attention & Use of physical artifacts & enhanced learning \\
& positive emotions \& behavioral traits & Work-out problems and examples & exam preparation and guidelines \\

& humor in teaching (examples \& instructor role) & & challenging course load \\
& general humor usage in instruction & & experiment explanation request \\
& insufficient detail & & engineering career preparation experiences \\
& & & homework quality \& support \\
& & & student engagement and learning environment \\
& & & grading and timeliness issues \\
& & & \\
\bottomrule
\end{tabular}}
\caption{EECS vs. Manual Code Comparison (Part 2)}

\end{table}

\begin{table}[!htb]
\centering
\resizebox{\textwidth}{!}{\begin{tabular}{llll}
\toprule
Instructional Resources & Instructional Resources (EECS) & Content and Sequence (EECS) & Drop Codes (EECS) \\
\midrule
Classroom technology & powerpoint presentation effectiveness and clarity & course structure \& organization & engineering major/courses selection \\
Demonstrate structures and processes to solve work-out problems & organizational structure & duration and utilization & lack of specific help \\
Instructor-provided notes & resource efficiency & & increased self-confidence \\
Online utilities & external support & & lack of specific feedback \\

Review sessions & technology for enhanced engagement & & layout confusion \\
Work-out problems and examples & equipment maintenance \& issues & & \\
& platform usefulness evaluation & & \\
& online platform types & & \\
& non-linear slide structure & & \\
& dyknow presentation design & & \\
& office hours effectiveness & & \\

& effective review sessions & & \\
& planning services & & \\
& reference materials (formula sheets) & & \\
& graduate teaching assistant role & & \\

\bottomrule
\end{tabular}}
\caption{EECS vs. Manual Code Comparison (Part 3)}
\end{table}

\end{landscape}
\restoregeometry

\newpage
\bibliographystyle{ieeetr}
\bibliography{all_refs}

\end{document}